\documentclass[10pt]{article}
\usepackage{amsmath}
\usepackage{amsfonts}
\usepackage{amssymb}
\usepackage[version=4]{mhchem}
\usepackage{stmaryrd}
\usepackage{graphicx}
\usepackage[export]{adjustbox}
\graphicspath{ {./images/} }
\usepackage{hyperref}
\hypersetup{
  colorlinks=true,
  linkcolor=blue,
  filecolor=magenta,
  urlcolor=cyan
}
\title{\bfseries Accelerating Object Detection with YOLOv4 for Real-Time Applications }
\vspace{1em}

\author{
       K. Senthil Kumar\textsuperscript{1},    K.M.B. Abdullah Safwan\textsuperscript{1}\\[1em]
    \textsuperscript{1}Department of Aerospace Engineering, Madras Institute of\\
    Technology, Anna University
}
\date{}

\begin{document}
\maketitle

\begin{abstract}
Object Detection is related to Computer Vision. Object detection enables detecting instances of objects in images and videos. Due to its increased utilization in surveillance, tracking system used in security and many others applications have propelled researchers to continuously derive more efficient and competitive algorithms. However, problems emerges while implementing it in real-time because of their dynamic environment and complex algorithms used in object detection. In the last few years, Convolution Neural Network (CNN) have emerged as a powerful tool for recognizing image content and in computer vision approach for most problems. In this paper, We revived begins the brief introduction of deep learning and object detection framework like Convolutional Neural Network(CNN), You only look once - version 4(YOLOv4). Then we focus on our proposed object detection architectures along with some modifications. The traditional model detects a small object in images. We have some modifications to the model. Our proposed method gives the correct result with accuracy.
\end{abstract}


\section*{1 Introduction}
Unmanned aerial vehicles (UAVs) have seen tremendous growth in recent years and are being applied across various fields like surveillance, fire detection, reconnaissance, mapping, and search and rescue missions. Many of these applications require real-time observation, but detecting objects through UAV imagery poses significant challenges. The images captured by these drones often suffer from poor quality, including noise and blur caused by the motion of the UAV. Furthermore, most onboard cameras have low resolution, making it difficult to spot small targets in the images.

The challenge becomes more intense due to the need for near real-time object detection in many UAV operations. Researchers have been working on ways to detect and track specific objects like vehicles, people, and landmarks for tasks such as autonomous navigation and safe landings. One promising solution is the use of convolutional neural networks (CNNs), a type of deep learning model designed for image processing. CNNs are particularly suited for image tasks because they make use of local receptive fields and shared weights, which allow them to detect patterns across an image without being influenced by the exact location of objects.

These CNN models are usually much deeper than traditional neural networks, often consisting of many layers, which enables them to process complex visual data. CNNs have proven to be highly effective in computer vision tasks, making them ideal for UAVs that need to perform object detection efficiently and accurately in realtime. By recognizing patterns and generalizing them across the entire image, CNNs play a crucial role in ensuring successful object detection for UAV missions, even in dynamic or difficult environments.

\subsection*{1.1 Object Detection}
Object detection is a computer technology related to computer vision and image processing that deals with detecting instances of semantic objects of a certain class such as human, buildings, or cars. Every object has its own special features that help in classifying the class. Object detection uses these uses these special features. For eg when looking for the circles object that are perpendicular distance from a centre point. When looking for the square, objects that are perpendicular at the corners, and have equal side lengths are needed.

\subsection*{1.2 Need For Object Detection}
The relevance of UAV plays an important role in places where human invention is impossible. Having accurate and up-to-date data from intelligence, surveillance and reconnaissance mission has become an essential part of how the modern traction develops strategy. To make UAV available for these application target detection is very much necessary and tracking to make an UAV fully autonomous during these applications. The accuracy of aerial data is directly related to the spatial resolution of the input imagery which is key operation to detect, locate the target. The advancement of computing practices resulted in vigorous and fully automatic production practices and coupled with hign-end computing machines and viewing machanisms will deal with positional accuracies and imagery orientation information which are characteristically challenging with customary techniques.

\subsection*{1.3 Proposed Idea}
Convolutional neural networks (convNets or CNNs) are a category of neural networks that have proven very effective in areas such as image recognition and classification. ConvNets are successful in identifying faces, objects, and traffic signs apart from powering vision in robots and self-driving cars. In dynamic environments, large visual variations occur due to changes in the background and lighting, hence they demand an advanced discriminative model to accurately differentiate targets from the backgrounds. Consequently, effective models for the problem tend to be computationally prohibitive. To address these two conflicting challenges, the proposed idea is using convolutional neural networks (CNNs), which have a very powerful discriminative capability, while maintaining high performance. Convolution neural networks operate at multiple resolutions, quickly reject the background regions in the fast low-resolution stages, and accurately evaluate trained targets in the dynamic environment.

CNNs are basically just several layers of convolutions with nonlinear activation functions like ReLU or tanh applied to the results. In a traditional feedforward neural network, we connect each input neuron to each output neuron in the next layer. That's also called a fully connected layer, or affine layer. In CNNs, we don't do that. Instead, we use convolutions over the input layer to compute the output. This results in local connections, where each region of the input is connected to a neuron in the output. Each layer applies different filters, typically hundreds or thousands like the ones shown above, and combines their results. There's also something called pooling (subsampling) layers, but I'll get into that later. During the training phase, a CNN automatically learns the values of its filters based on the task you want to perform. For example, in Image Classification, a CNN may learn to detect edges from raw pixels in the first layer, then use the edges to detect simple shapes in the second layer, and then use these shapes to detect higher-level features, such as facial shapes in higher layers. The last layer is then a classifier that uses these high-level features.

Moving object tracking is one of the challenging problems in computer vision and it has numerous applications in surveillance systems, traffic monitoring, etc. The goal of the object tracking algorithm is to locate a moving object in consecutive video frames. Tracking a moving object in a video becomes difficult due to the random motion of objects. Even though many algorithms have been developed and many applications of object tracking have been made, object tracking is still considered a difficult task to accomplish. The existence of several problems such as illumination variation, tracking non-rigid objects, non-linear motion, occlusion, and the requirement of real-time implementation has made tracking one of the most challenging tasks in computer vision. The problem of tracking a moving object in a dynamic background is divided into three parts: object detection using an image processing algorithm, estimation of the object motion, and compensation of the camera motion. To counterattack the difficulties in detection, a convolutional neural network is used, and the coordinates from the detection are passed to the tracker, and input to the tracker is updated regularly from the detection algorithm.

\newpage
\subsection*{1.4 Working Process For Detecting The Images}
\begin{itemize}
  \item Darknet Setup.
  \item Labeled Custom Dataset.
  \item Python code to detect the object using YOLOV4 algorithms.
  \item Train the detector.
  \item Check performance.
  \item Test your custom Object Detector.
\end{itemize}

\section*{2 Working of YOLOv4 Algorithm}
YOLOv4 is known for its up-gradation in terms of AP and FPS. YOLOv4 prioritizes real-time object detection, and training takes place on a single CPU. YOLOv4 has obtained state-of-the-art results on the COCO dataset with 43.5\% speed (AP) at 65 FPS (Performance) on Tesla V100. This achievement is the result of a combination of features like DropBlock Regularization, Data Augmentation, Mish-Activation, CrossStage Partial connections (CSP), Self adversarial training (SAT), Weighted-Residual-Connections (WRC), and many more. 

There are two types of models: one-stage and two-stage object detectors. In two-stage detectors, detection works in two parts: first, regions of interest are detected, and then regions are classified to determine if the object is present. YOLOv4, being a single-stage object detector, is both more accurate and faster than two-stage detectors like R-CNN and Fast R-CNN.\\
\includegraphics[max width=\textwidth, center]{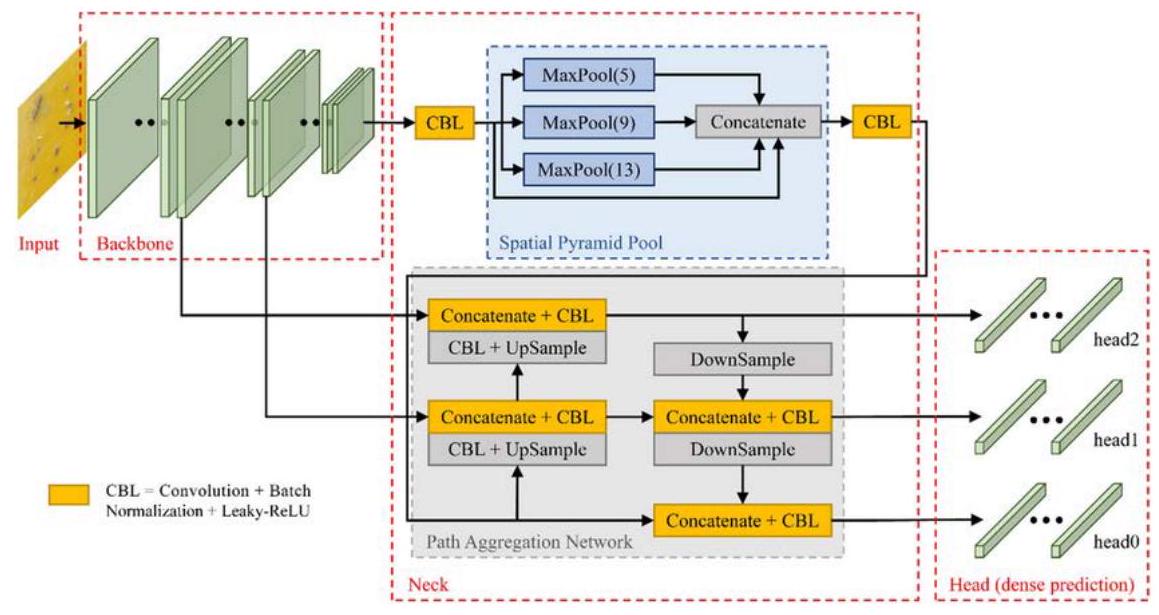}

\textbf{Figure 1:} Basic YOLOv4 Architecture

\subsection*{2.1 Backbone}
YOLOv4 uses one of three models as its backbone. These feature extractor models include:

\subsubsection*{2.1.1 CSPResNext50}
Both CSPResNext50 and CSPDarknet53 are DenseNet-based models. CSPResNext50 works similarly to CSPDarknet53 and operates on the CSPNet strategy. When considering the COCO dataset, CSPDarknet53 outperforms CSPResNext50 in classifying objects. CSPResNext50 consists of 16 CNN layers with a receptive field of $425 \times 425$ and 20.6 million parameters, while CSPDarknet53 consists of 29 CNN layers with a receptive field of $725 \times 725$ and 27.6 million parameters.

\subsubsection*{2.1.2 CSPDarknet53}
This is a widely used backbone for object detection, making use of DarkNet-53. YOLOv4 specifically uses CSPDarknet53 as its backbone. It operates on the CSPNet strategy, dividing the DenseBlock feature map into two halves and merging them through a cross-stage hierarchy. The first part bypasses the base layer and is used as input for the next transition layer, while the second part undergoes DenseBlock. This strategy reduces computational complexity. CSPDarknet53 achieves higher accuracy than other ResNet models and offers excellent performance.\\
\includegraphics[max width=\textwidth, center]{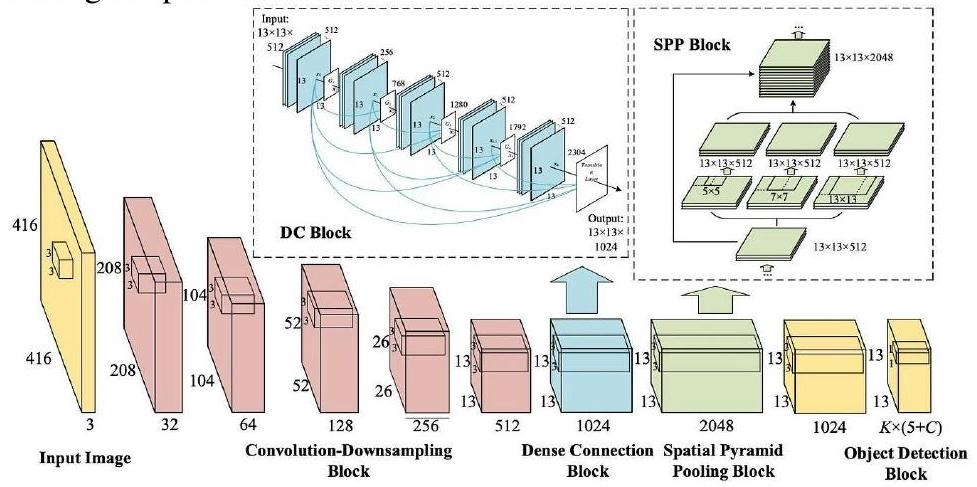}

\textbf{Figure 2:} CSPDarknet53 Architecture

\subsubsection*{2.1.3 EfficientNet-B3}
EfficientNet-B3 is an image classification model used to attain state-of-the-art accuracy. It restudies convolutional neural network scaling and is based on AutoML. The AutoML mobile framework was developed to create a small-sized network known as EfficientNet-B0. Compound Scaling, as the name suggests, helps scale up the AutoML baseline, resulting in EfficientNet-B1 through EfficientNet-B7.

\subsection*{2.2 Neck}
The second stage in the YOLOv4 pipeline, the "Neck," gathers features formed in the backbone and feeds them to the head for detection. YOLOv4 provides several options for this stage:

\subsubsection*{2.2.1 FPN (Feature Pyramid Networks)}
Earlier detectors predicted objects based on a pyramidal feature hierarchy extracted from the backbone. To address the issue of effective representation and multi-scale feature processing, FPN was introduced, which gathers features from different scales through a top-down path.\\
\includegraphics[max width=\textwidth, center]{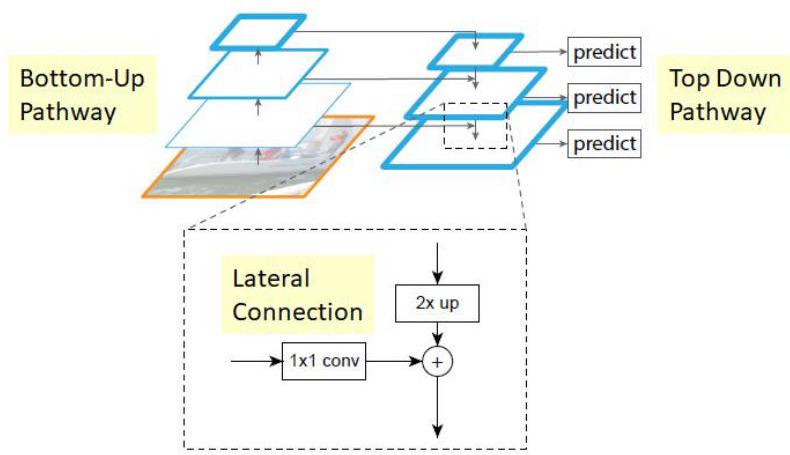}

\textbf{Figure 3:} FPN (Feature Pyramid Networks)

\subsubsection*{2.2.2 PAN (Path Aggregation Network)}
PAN is used as a neck in the YOLOv4 algorithm to enhance the process of segmentation by maintaining semantic information, helping accurately localize image elements for mask creation.\\
\includegraphics[max width=\textwidth, center]{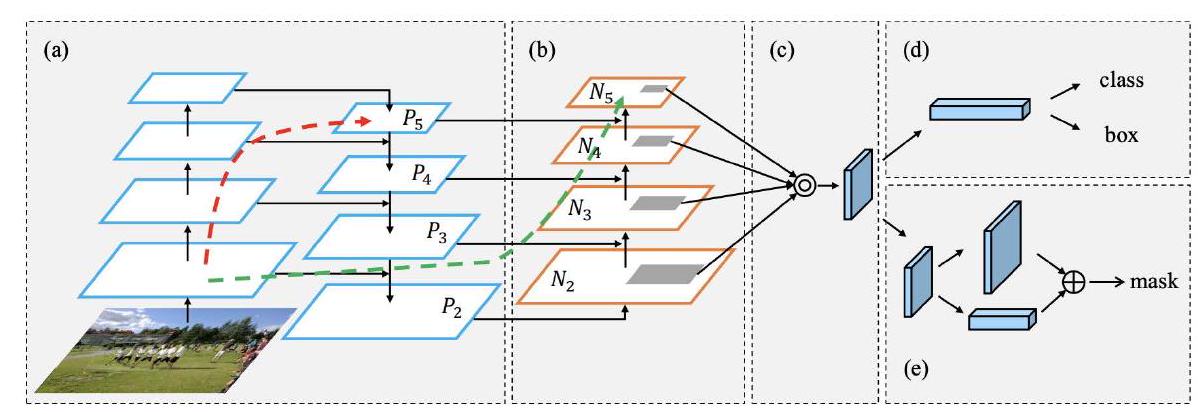}

\textbf{Figure 4:} PANet Architecture

\subsubsection*{2.2.3 Properties of PAN}
When an image passes through various layers of a neural network, its spatial resolution decreases while the complexity of its features increases. This makes pixel-level mask identification challenging. In YOLOv4, FPN uses a top-down approach to combine multi-scale features and preserve semantic localization information. However, producing masks for large objects can become intricate because information must pass through multiple layers. PANet adds a bottom-up and top-down route to alleviate this complexity by using lateral connections from lower to upper layers. It consists of at most 10 layers, making it less complicated and more efficient than FPN.

\subsubsection*{2.2.4 Adaptive Feature Pooling}
Earlier techniques like Mask-RCNN relied on features from a single stage for prediction. However, the use of RIO Align Pooling to extract features from upper levels sometimes produced inaccurate results. PANet addresses this issue by extracting features from all layers, performing Align Pooling on each feature map to extract the relevant features.

\subsubsection*{2.2.5 Fully-Connected Fusion}
In Mask-RCNN, a fully convolutional network is used to conserve spatial information while reducing the number of parameters. However, this model cannot learn how to use pixel locations for prediction, as the parameters are shared across all spatial positions. Fully connected layers, being location-sensitive, can adjust to different locations. PANet uses data from both layers to ensure precise prediction.

\subsubsection*{2.2.6 SPP (Spatial Pyramid Pooling)}
SPP, or Spatial Pyramid Pooling, is used to obtain both fine and coarse information. It applies pooling using kernels of various sizes ($1 \times 1$, $5 \times 5$, $9 \times 9$, $13 \times 13$). These maps are combined to produce the final output. The advantage of using SPP is the improved receptive field. It creates fixed-size features regardless of the feature map size, making it robust to object deformations. SPP is also flexible in terms of input scales, as it can extract pooled features at variable scales.\\
\includegraphics[max width=\textwidth, center]{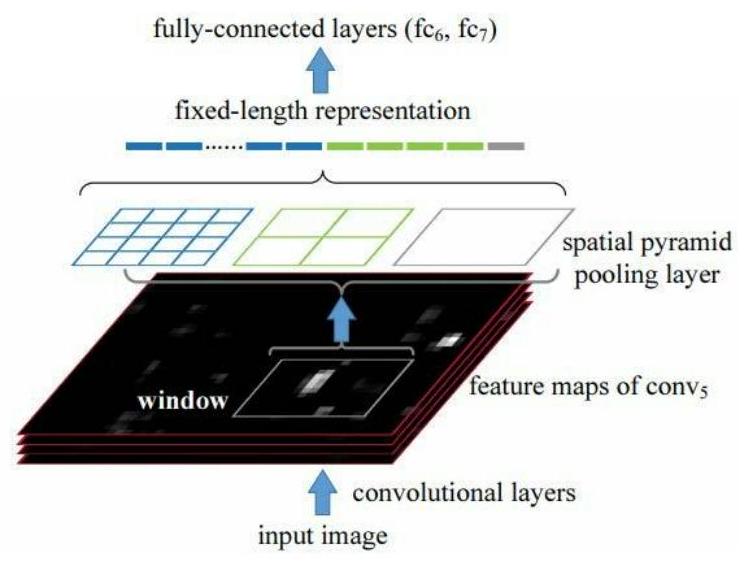}

\textbf{Figure 5:} Spatial Pyramid Pooling

\subsection*{2.3 Head}
The main objective of the head in YOLOv4 is to perform predictions, including both classification and regression of bounding boxes. The YOLOv4 head is responsible for determining the precise location and dimensions of objects within an image while simultaneously predicting the class of each object. This unified approach, unlike models that separate these tasks, improves both speed and accuracy, making YOLOv4 particularly well-suited for real-time applications. The head provides important information about the coordinates of bounding boxes (x, y, height $\mathrm{h}$, and width $\mathrm{w}$), which define the position and size of detected objects. In addition, the head outputs a confidence score and the predicted label for each object.

YOLOv4 head builds upon the YOLOv3 architecture but introduces optimizations for better performance. Each anchor box in the YOLOv4 head is analyzed for potential object detection, and if an object is found, the model refines the bounding box. The key feature of YOLOv4 is its balance between speed and computational efficiency, making it suitable for real-time applications like autonomous driving, video surveillance, and drone navigation. Despite its advanced capabilities, it can be deployed on standard hardware without significantly increasing computational demands.\\
\includegraphics[max width=\textwidth, center]{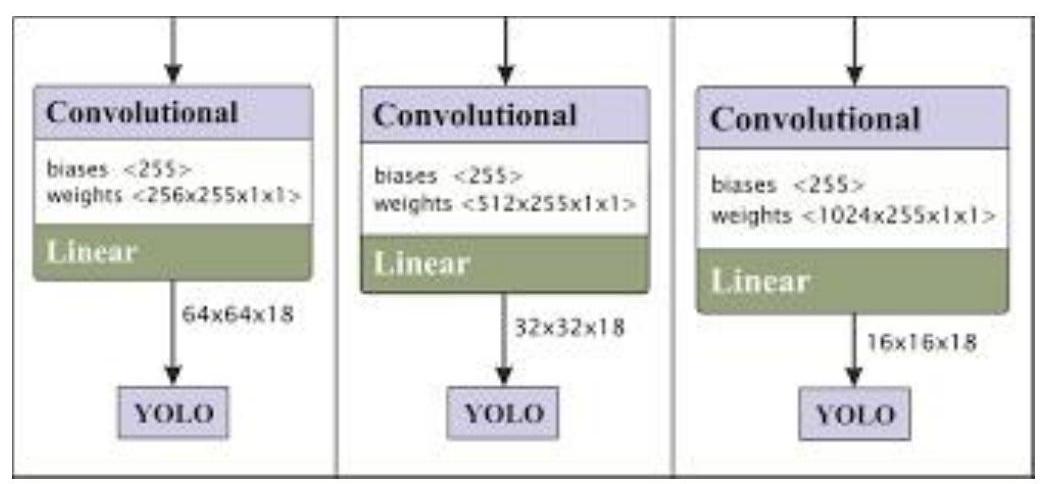}

\textbf{Figure 6:} YOLOv4 head applied at different scales

\subsection*{2.4 Anchor Box}
Anchor boxes are used to detect multiple objects of varying sizes in a single frame, all centered in the same cell. This contrasts with traditional grids that detected a single object per frame. If the number of anchor boxes changes, the length of the ground truth and prediction arrays also changes.

For example, consider a box in the cell with 80 prediction classes, i.e., [Pc, P1, P2...P80, X1, Y1, X2, Y2], which totals 85 values. For these, 9 anchor boxes will be used, resulting in an array length of $85 \times 9 = 765$ predictions.

\subsection*{2.5 Bag of Freebies in YOLOv4}
A "Bag of Freebies" refers to techniques that improve the performance of a model during training, without increasing the inference time during deployment. These techniques enhance model accuracy through data management and augmentation strategies. They enrich the dataset by exposing models to diverse situations that the model may not have encountered otherwise. Data augmentation strategies help generalize the model and improve robustness, allowing it to learn from a wider variety of situations.

\subsection*{2.6 Bag of Specials in YOLOv4}
The "Bag of Specials" consists of techniques and models that enhance post-processing and coherence at a minor increase in inference cost. However, these techniques improve the detector's overall accuracy and performance. The selection of specific techniques depends on the architecture and technical parameters of the model, but ultimately, the goal is to refine and improve the overall model.

\section*{3 Tools Required for Detection and Estimation}
\subsection*{3.1 Software Requirements}
\begin{itemize}
  \item Python
  \item Google Colab
  \item Darknet
  \item YOLOv4
\end{itemize}

\subsubsection*{3.1.1 Python}
Python is an open-source programming language that simplifies solving computational problems. It allows the code to be written once and executed across different platforms without modification. Python is an interpreted language, meaning it converts human-readable code into machine-executable instructions at runtime. Traditionally, interpreted languages were referred to as scripting languages, implying their use for trivial tasks. However, modern languages like Python are now used for large-scale applications.

\subsubsection*{3.1.2 Google Colab}
Google Colab allows users to write and execute Python code directly in the browser, with zero configuration required. It provides free access to GPUs and facilitates easy sharing of code. Colab is especially suited for machine learning, data analysis, and educational purposes. Colab notebooks combine executable code, text, images, HTML, and LaTeX in a single document. These notebooks are stored in Google Drive, enabling easy sharing and collaboration with others.

\begin{itemize}
  \item Zero configuration required
  \item Free access to GPUs
  \item Easy sharing
\end{itemize}

\subsubsection*{3.1.3 Darknet}
A darknet is an overlay network on the Internet that requires special software, configurations, or authorizations to access. It often uses customized communication protocols. Two main types of darknets include social networks (for peer-to-peer file sharing) and anonymity proxy networks such as Tor. Technology like Tor, I2P, and Freenet was originally designed to protect digital rights by ensuring security, anonymity, and resistance to censorship. These networks are used for both legal and illegal activities, facilitating anonymous communication between whistleblowers, activists, journalists, and news organizations through tools such as SecureDrop.

\subsubsection*{3.1.4 YOLOv4}
YOLOv4 introduces improvements in Average Precision (AP) and Frames Per Second (FPS). It prioritizes real-time object detection and can train models using a single CPU. On the COCO dataset, YOLOv4 achieved state-of-the-art results with $43.5\%$ AP at 65 FPS on Tesla V100. These improvements are attributed to features such as DropBlock Regularization, Data Augmentation, Mish Activation, Cross-Stage Partial Connections (CSP), Self-Adversarial Training (SAT), and Weighted Residual Connections (WRC), among others.

\section*{4 Train YOLOv4 Custom Object Detector}
We will now create a custom YOLOv4 object detector to recognize specific classes or objects. The following steps and tools are required:

\begin{itemize}
  \item Darknet setup
  \item Labeled custom dataset
  \item Custom .cfg file
  \item obj.data and obj.names files
  \item train.txt file (test.txt is optional)
\end{itemize}

\subsection*{4.1 Darknet Setup}
The first step is to clone Darknet from AlexeyAB's GitHub repository. After cloning, adjust the `Makefile` to enable OpenCV and GPU support for Darknet. Once the adjustments are made, build Darknet using the `make` command in Google Colab. This repository contains configuration files required to perform object detection using the YOLOv4 algorithm.

\subsection*{4.2 Gathering and Labeling a Custom Dataset}
To create a custom object detector, you need a comprehensive dataset of images and labels that the detector can use to train. There are two main ways to create this dataset: using Google images or creating your own dataset, followed by using an annotation tool to manually label the images.

\subsubsection*{4.2.1 Method 1: Using Google's Open Images Dataset}
This method is highly recommended, as it allows you to gather thousands of images with auto-generated labels in just a few minutes. Google's Open Images Dataset, in combination with the OIDv4 toolkit, provides a fast and efficient way to generate labeled datasets. The dataset contains labeled images across more than 600 classes.

\paragraph{4.2.1.1 Training Dataset} \mbox{}\\[1em]
The following commands were used within the toolkit to generate the custom training dataset. For optimal accuracy, I chose to use 1500 images, but as a general rule, the more images you use, the higher the accuracy of your model.

\paragraph{4.2.1.2 Validation Dataset} \mbox{}\\[1em]
A validation dataset is essential for testing the custom object detector after training. The OIDv4 Toolkit provides separate images for validation to ensure there is no overlap between training and validation datasets. You can run the same commands as used for the training dataset, but for validation purposes. It is recommended that the validation dataset be around $20-30\%$ of the size of your training dataset. For example, if 1500 images were used for training, the validation dataset should contain about 300 images. $(20\%$ of $1500 = 300)$.

\paragraph{4.2.1.3 Converting Labels to YOLOv4 Format} \mbox{}\\[1em]
The labels generated by the OIDv4 Toolkit are not in YOLOv4 format by default. However, they can be easily converted with a few simple commands. First, open the `classes.txt` file within the root OIDv4\_Toolkit folder, and edit the file to include the classes you just downloaded, one class per line.

Next, run the \texttt{convert\_annotations} Python file to convert the OIDv4 labels into YOLOv4 format. This process converts the labels for both the training and validation datasets.

Once the labels are converted, they can be used by Darknet to train the custom object detector. Be sure to remove the old `Label` folder from both the train and validation directories, as it contains non-YOLOv4 formatted labels. Then, delete the old labels for the validation set as well.

\begin{verbatim}
LABEL FORMAT
  *2a885eb4d636684f - Notepad
File Edit Format View Help
Tank 165.000192 200.300544 290.69952 287.09376
Truck 82.100224 163.443949 912.346112 556.2229060000001

YOLOv4 FORMAT
  *007cae31ad106eac - Notepad
File Edit Format View Help
0 0.4087525274151436 0.6863815 0.7662260000000001 0.579757
1 0.49315701044386423 0.2931798125 0.767878 0.542647
\end{verbatim}

\textbf{Figure 7:} YOLOv4 Label Format

\subsubsection*{4.2.2 Method 2: Manually Labeling Images with Annotation Tool}
If you are unable to find the necessary images or classes in Google's Open Images Dataset, you will need to manually label your images using an annotation tool. This can be a time-consuming process. In a previous tutorial, I demonstrated how to mass-download images from Google Images and use LabelImg, an annotation tool, to create a custom dataset for YOLOv4. After following these steps, you should have a folder containing images and text files for both your training and validation datasets, as shown below.\\
\includegraphics[max width=\textwidth, center]{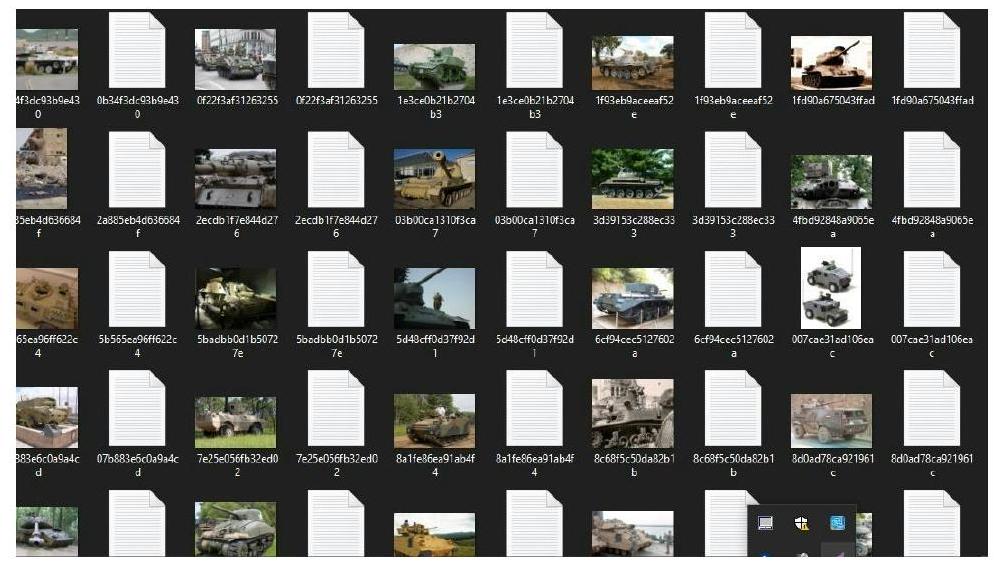}

\noindent\textbf{Figure 8:} Custom YOLOv4 Training and Validation Datasets with Proper Labels

\subsection*{4.3 Moving Your Custom Datasets Into Your Cloud VM}
Once your datasets are properly formatted for training and validation, you need to move them into your cloud VM. I recommend renaming the folder containing your training images and text files to `obj`, then compressing the folder into a `.zip` file called `obj.zip`. Next, upload the `.zip` file to your Google Drive. Do the same for the validation dataset, naming it `test`. You should now have both `obj.zip` and `test.zip` in your Google Drive. This will significantly reduce the time required to transfer your dataset to the cloud VM. Once uploaded, you can copy and unzip the files in your cloud VM.

\subsection*{4.4 Configuring Files for Training}
The next step is configuring the necessary files for training: the custom `.cfg` file, `obj.data`, `obj.names`, `train.txt`, and `test.txt`. Careful attention is required when editing these files, as even minor errors can cause significant issues during training.

\subsubsection*{4.4.1 Configuration File (Cfg File)}
The `yolov4.cfg` file must be copied to your Google Drive for editing. Open the file in a text editor and make the necessary adjustments to match your custom object detector's needs. You can edit the configuration using Google Drive's built-in Text Editor by double-clicking `yolov4-obj.cfg` and selecting "Open with Text Editor" from the dropdown menu.

For optimal results, set `batch=64` and `subdivisions=16`. If you encounter any issues, increase `subdivisions` to 32. Additional changes depend on how many classes you are training your detector on. Set `max\_batches` to $6000$, and `steps` to $4800$, $5400$. Change the `classes` parameter to `1` in the three YOLO layers, and adjust the `filters` parameter to 18 in the three convolutional layers preceding the YOLO layers.

To configure the network size, set `width=416` and `height=416` (or any multiple of 32). The `max\_batches` value should be calculated as 
\[
\text{Number of Classes} \\ \times 2000
\]
but it should not be less than 6000. For example, if training 1-3 classes, `max\_batches` should be set to 6000, but for 5 classes, it should be set to 10000. The `steps` should be set to $80\%$ and $90\%$ of `max\_batches`. For example, if `max\_batches = 10000`, then `steps` should be set to 8000 and 9000. For `filters`, calculate the value as 
\[
(\text{Number of Classes} + 5) \times 3
\]
For example, if training 1 class, set `filters=18`; if training 4 classes, set `filters=27`.

Optional: If you experience memory issues or find the training process too slow, change the line `random=1` to `random=0` in each of the three YOLO layers. This will reduce training time and memory usage, but it will slightly decrease accuracy.

\subsubsection*{4.4.2 obj.names and obj.data Files}
Create a new file within a code or text editor called \texttt{obj.names}, where you will have one class name per line in the same order as your \texttt{classes.txt} from the dataset generation step. You will also create an \texttt{obj.data} file and fill it in like this (change the number of classes accordingly, as well as your backup location). This backup path is where we will save the weights of our model throughout training. Create a backup folder in your Google Drive and put its correct path in this file.\\
\includegraphics[max width=\textwidth, center]{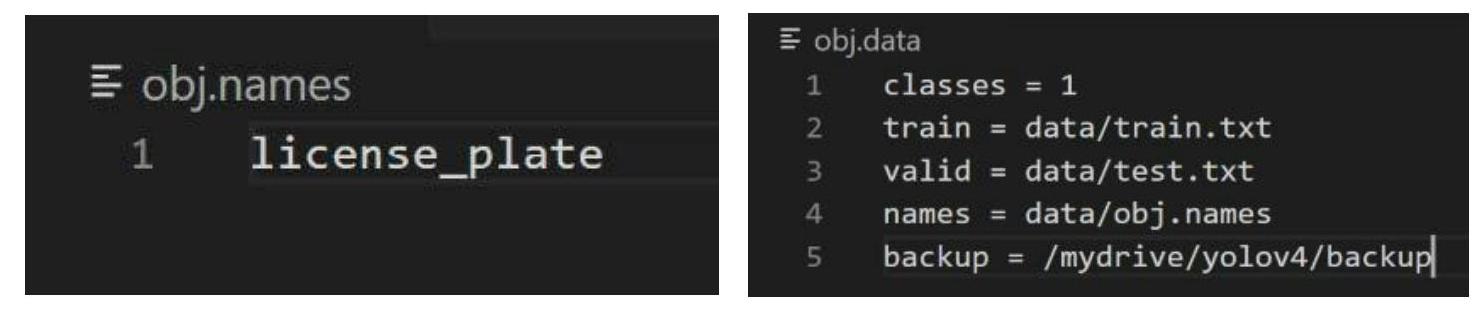}

\textbf{Figure 9:} obj.names and obj.data text files

\subsubsection*{4.4.3 Generating train.txt and test.txt}
The last configuration files needed before we can begin to train our custom detector are the \texttt{train.txt} and \texttt{test.txt} files, which hold the relative paths to all our training images and validation images. We have scripts that easily generate these two files with the proper paths to all images. The scripts can be accessed from the GitHub Repo. Just download the two files to your local machine and upload them to your Google Drive so we can use them in the Colab Notebook. Now simply run both scripts to generate the two text files. If you are uncertain whether generating the files worked and want to double-check that everything went as planned, double-click on \texttt{train.txt} on the left side of the File Explorer, and it should contain one line for each training image path, like in the figure below.

\begin{verbatim}
train.txt
data/obj/46bffa67f6381212.jpg
data/obj/76b61bd5960a322f.jpg
data/obj/7bdee48d3c08ecb1.jpg
dala/ubj/bu68c499ba0b5f85.jpg
data/obj/fcfe48a6bca3e713.jpg
data/obj/2c075189b105430b.jpg
data/obj/4a436c1daff4719c.jpg
data/obj/b1f86fcb2b9dd575.jpg
data/obj/c12d150c0d965d0a.jpg
\end{verbatim}

\textbf{Figure 10:} train.txt file

\subsection*{4.5 Download Pre-trained Weights for the Convolutional Layers}
This step downloads the weights for the convolutional layers of the YOLOv4 network. By using these weights, your custom object detector will be more accurate, and the training time will be reduced. You do not have to use these weights, but they will help your model converge faster and become more accurate.

\texttt{!wget https://github.com/AlexeyAB/darknet/releases/download/} \\
\texttt{darknet\_yolo\_v3\_optimal/yolov4.conv.137}

Use the above command to download the pre-trained weight file in Colab.

\subsection*{4.6 Train Custom Object Detector}
The time has finally come. You have made it to the moment of truth. You are now ready to train your custom YOLOv4 object detector on whatever custom classes you have decided on. Run the following command. The \texttt{-dont\_show} flag stops the chart from popping up since Colab Notebook can't open images on the spot, and the \texttt{-map} flag overlays mean average precision on the chart to see the accuracy of your model. Only add the \texttt{-map} flag if you have a validation dataset.\\
\texttt{!./darknet detector train <path to obj.data> <path to custom}\\
\texttt{config> yolov4.conv.137 -dont\_show -map}

Run the above code in Google Colab to train the custom object detector by providing the path to the training and validation dataset. After training, you can observe a chart of how your model performed throughout the training process by running the command below. It shows a chart of your average loss vs. iterations. For your model to be considered 'accurate,' you should aim for a loss under 2.\\
\includegraphics[max width=\textwidth, center]{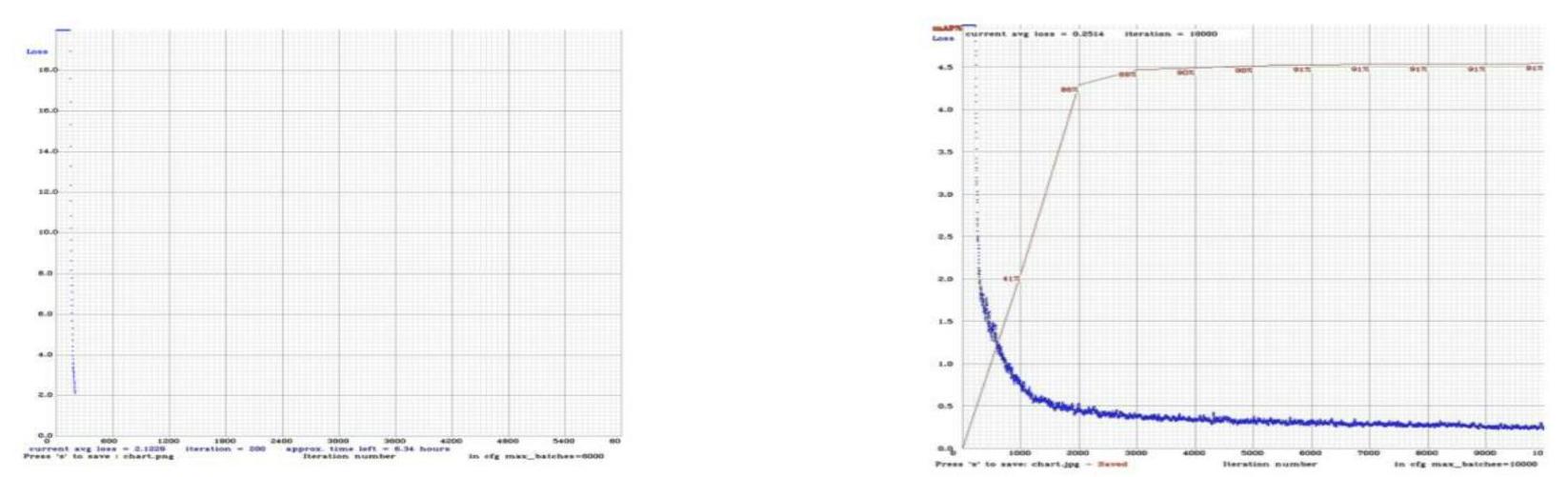}

\textbf{Figure 11:} Model Loss at 500th iteration and 10000th iteration.
\vspace{1em}

If for some reason you get an error or your Colab goes idle during training, you have not lost your partially trained model and weights! Every 100 iterations, a weights file called \texttt{yolov4-obj\_last.weights} is saved to \texttt{mydrive/yolov4/}\\
 \texttt{backup/}(or wherever your backup folder is). This is why we created this folder in our Google Drive and not on the cloud VM. If your runtime crashes and your backup folder was in your cloud VM, you would lose your weights and your training progress.

\texttt{!./darknet detector train data/obj.data cfg/yolov4-obj.cfg}\\
\texttt{/mydrive/yolov4/backup/yolov4-obj\_last.weights -dont\_show}

We can kick off training from our last saved weights file so that we don't have to restart. Just run the above command with your backup location.

\subsection*{4.7 Checking the Mean Average Precision (mAP) of our Model}
If you didn't run the training with the \texttt{-map} flag added, you can still find out the mAP of your model after training. Run the following command on any of the saved weights from the training to see the mAP value for that specific weight file. I suggest running it on multiple saved weights to compare and find the weights with the highest mAP, as that will be the most accurate one.\\
\texttt{!./darknet detector map data/obj.data cfg/yolov4-obj.cfg /mydrive}\\
\texttt{/yolov4/backup/yolov4-obj\_1000.weights}

\textbf{NOTE:} If you think your final weights file has overfitted, it is important to run these mAP commands to see if one of the previously saved weights is a more accurate model for your classes.

\subsection*{4.8 Run Custom Object Detector}
We now have a custom object detector to make your very own detections. Time to test it out.\\
\texttt{!./darknet detector test data/obj.data cfg/yolov4-obj.cfg/mydrive}\\
\texttt{/yolov4/backup/yolov4-obj\_last.weights /mydrive/images/tanker.jpg}\\
\texttt{-thresh 0.3}\\
\texttt{imShow('predictions.jpg')}\\
To test the object detector, use the above command and provide the input image path. It will generate the detection for the input image and return the output.

\section*{5 Results \& Outputs}
\includegraphics[max width=\textwidth, center]{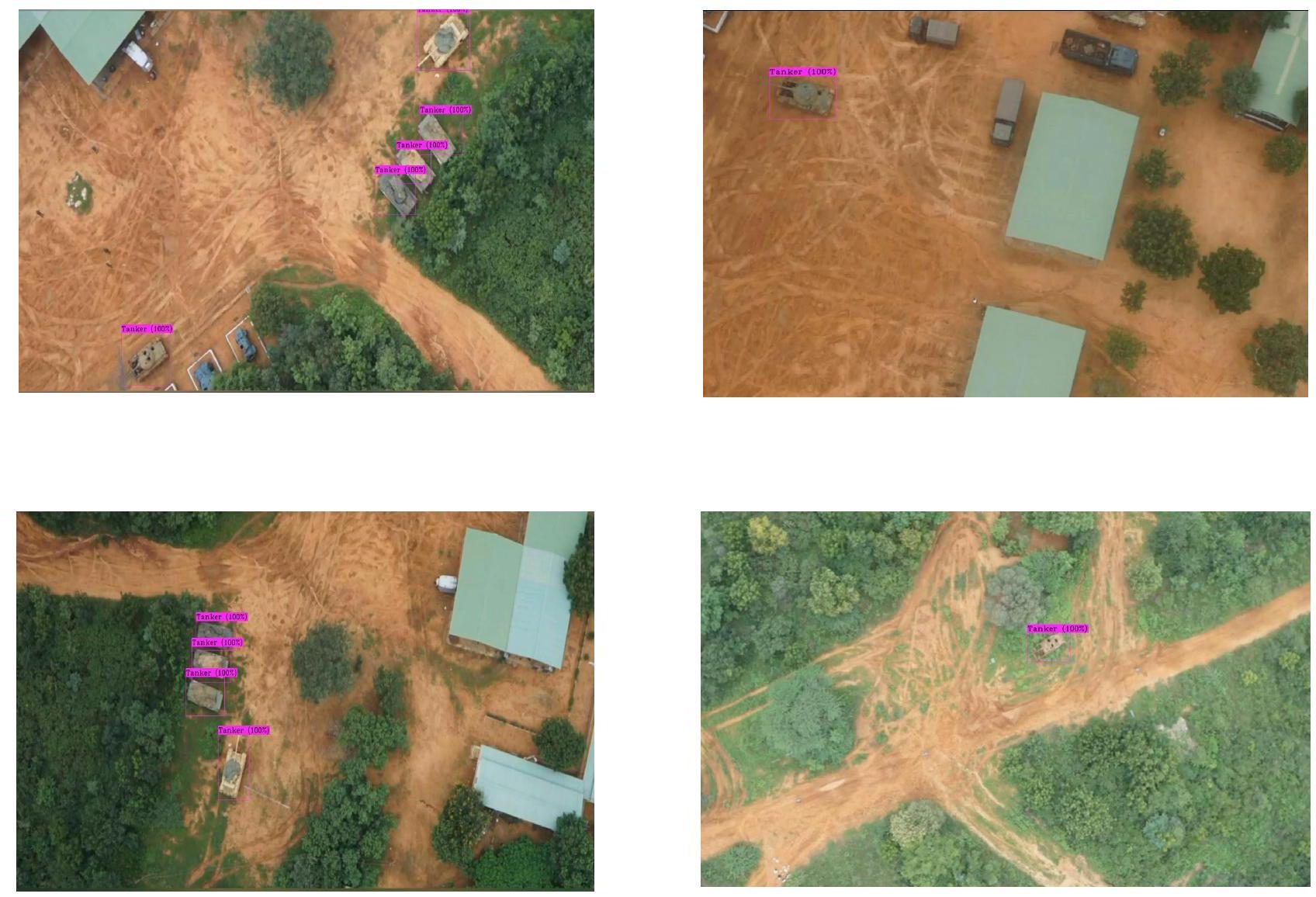}

\noindent\textbf{Figure 12:} Results \& Outputs of custom YOLOv4 detector on tanker

\section*{6 Conclusions}
In this project, we successfully implemented a real-time object detection system using the YOLOv4 algorithm. Through careful setup and configuration, we trained a custom object detector capable of recognizing specific objects efficiently. The YOLOv4 algorithm, known for its balance between speed and accuracy, allowed us to achieve state-of-the-art results while maintaining the flexibility to adapt to various datasets and object classes. We began by setting up the Darknet framework and preparing datasets, either by utilizing Google's Open Images Dataset or by manually labeling images. Next, we customized the configuration files, including \texttt{yolov4.cfg}, \texttt{obj.names}, and \texttt{obj.data}, tailoring them to meet our object detection needs. Additionally, we incorporated pre-trained weights to accelerate the training process, which significantly improved accuracy and reduced the overall training time.

Throughout the training process, we monitored key metrics such as loss values and mean average precision (mAP) to ensure the detector was learning effectively. In the event of interruptions, we safeguarded our progress by regularly saving weight files, allowing us to resume training without starting from scratch. Once trained, we tested the custom object detector on sample images, confirming that YOLOv4 delivers accurate and fast detections with minimal computational overhead. This project highlights the effectiveness of YOLOv4 for real-time object detection tasks, making it a strong candidate for applications in fields such as surveillance, autonomous driving, and industrial automation.

\end{document}